\documentclass[runningheads]{llncs}
\usepackage{graphicx}

\usepackage{amsmath,amssymb}
\usepackage{tikz}

\usepackage{orcidlink}
\usepackage{booktabs}
\usepackage{comment}
\usepackage{soul}
\usepackage{color}
\usepackage{cite}
\usepackage{multirow}
\usepackage{wrapfig}
\usepackage{fancyhdr}
\sethlcolor{green}

\usepackage[accsupp]{axessibility}  % Improves PDF readability for those with disabilities.

\fancypagestyle{disclaimer}{\fancyhf{}\fancyfoot[L]{
\footnotesize
This preprint has not undergone any post-submission improvements or corrections. The Version of Record of this contribution is published in Proceedings of the 17th European Conference on Computer Vision (ECCV 2022).
}}

\begin{document}
%% begin comment
% \linenumbers
%% end comment
\pagestyle{headings}
\mainmatter
\def\ECCVSubNumber{5293}  % Insert your submission number here

\title{DeepMend: Learning Occupancy Functions to Represent Shape for Repair} % Replace with your title

% \begin{comment}
% \titlerunning{ECCV-22 submission ID \ECCVSubNumber} 
% \authorrunning{ECCV-22 submission ID \ECCVSubNumber} 
% \author{Anonymous ECCV submission}
% \institute{Paper ID \ECCVSubNumber}
% \end{comment}
%******************

% \begin{comment}
\titlerunning{DeepMend: Learning Occupancy Functions to Represent Shape for Repair}
\author{Nikolas Lamb\orcidlink{0000-0002-6000-4658} \and
Sean Banerjee\orcidlink{0000-0003-3085-056X} \and
Natasha Kholgade Banerjee\orcidlink{0000-0001-7730-7754}\index{Banerjee, Natasha Kholgade}
}
\authorrunning{N. Lamb et al.}
\institute{
Clarkson University, Potsdam, NY 13699, USA\\ % TODO: check this
\email{\{lambne,sbanerje,nbanerje\}@clarkson.edu}
}
% \end{comment}
%******************

\maketitle

\begin{abstract}
We present DeepMend, a novel approach to reconstruct resto- rations to fractured shapes using learned occupancy functions. Existing shape repair approaches predict low-resolution voxelized restorations, or require symmetries or access to a pre-existing complete oracle. We represent the occupancy of a fractured shape as the conjunction of the occupancy of an underlying complete shape and the fracture surface, which we model as functions of latent codes using neural networks. Given occupancy samples from an input fractured shape, we estimate latent codes using an inference loss augmented with novel penalty terms that avoid empty or voluminous restorations. We use inferred codes to reconstruct the restoration shape. We show results with simulated fractures on synthetic and real-world scanned objects, and with scanned real fractured mugs. Compared to the existing voxel approach and two baseline methods, our work shows state-of-the-art results in accuracy and avoiding restoration artifacts over non-fracture regions of the fractured shape.

\thispagestyle{disclaimer}

\keywords{Learned Occupancy, Shape Representation, Repair, Fracture, Implicit Surface, Neural Networks}
\end{abstract}

\vspace{-3mm}

\begin{figure}[t]
    \centering
      \includegraphics[width=.9\linewidth]{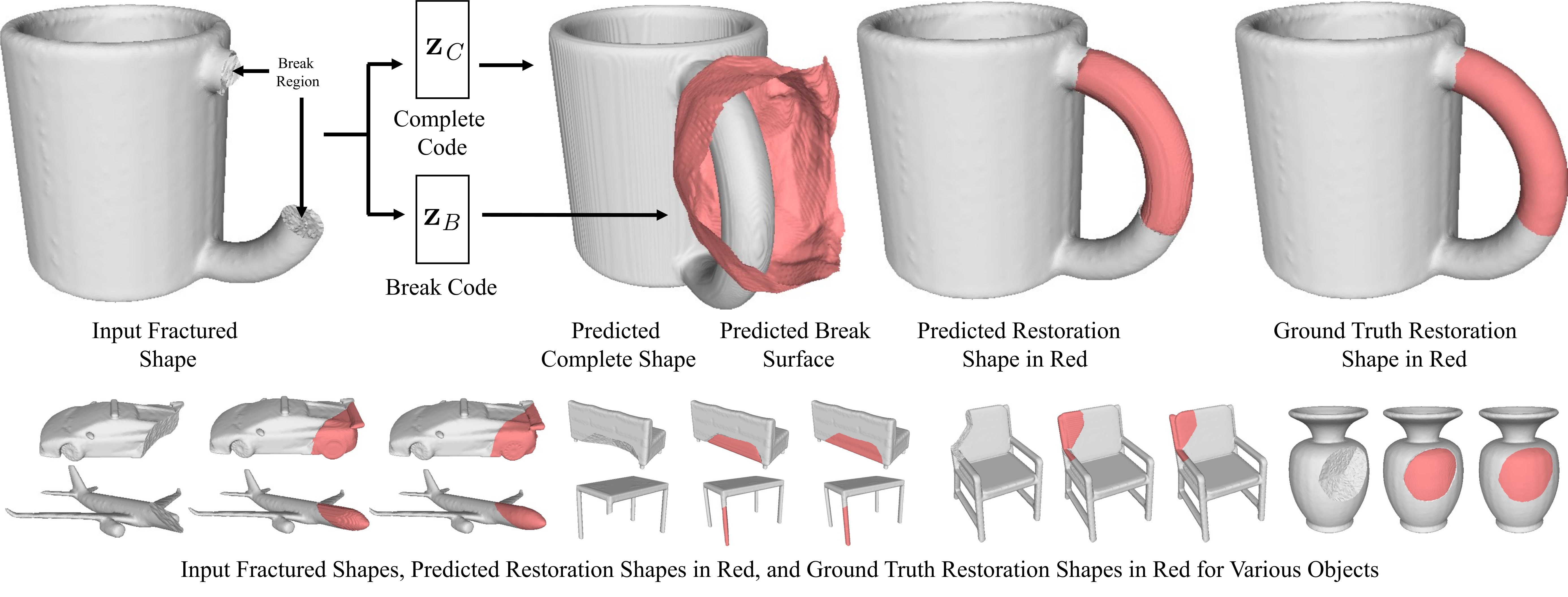}
      \caption{Given a fractured shape, our approach infers latent codes for an underlying complete shape and a break surface. We use the codes to generate a restoration shape that repairs the input fracture shape.}
    \label{fig:teaser}
\end{figure}

\section{Introduction}
Automated restoration of fractured shapes is an important area of study, with applications in consumer waste reduction, commercial recycling, cultural heritage object restoration, medical fields such as orthopedics and dentistry, and robot-driven repair. Despite its wide application, automated repair of fractured shapes has received little attention. Most current automated techniques use symmetries to complete fractured shapes~\cite{papaioannou2017reassembly, gregor2014towards}. These techniques do not generalize to objects with non-symmetrical damage. The only existing generalizable approach for repair operates in voxel space~\cite{hermoza20183d} and produces low-fidelity restorations. 

In this work, we present DeepMend, a novel deep learning approach to generate high-fidelity restoration shapes given an input fractured shape. Our approach is inspired by 
work on learning functions for quantities such as signed distance field (SDF) or occupancy that implicitly represent the shape surface over the continuous 3D domain~\cite{park2019deepsdf, sitzmann2020metasdf, tretschk2020patchnets, zheng2021deep, duggal2022mending, hao2020dualsdf, chen2019learning, yan2022shapeformer, tretschk2020patchnets}. These approaches perform partial shape completion by using the learned function to infer a latent code using field value samples from a partial shape, and to compute the complete shape values from the latent code. Different from partial shape completion, DeepMend addresses the challenge that, unlike a partial shape that is a subset of a complete shape, the fractured shape contains a novel break region missing in a complete shape, as shown in Figure~\ref{fig:teaser}. To restore fractured shapes, our main contribution is a novel representation that represents the occupancy of a fractured shape as the logical conjunction of occupancy values for a complete shape and a break surface. We use T-norms~\cite{gupta1991theory} to relax the logical conjunction into arithmetic operations. We represent the complete and break surface occupancy as functions parametrized on latent codes and modeled using deep neural networks.

Given an input fractured shape, we use the learned functions to automatically estimate latent codes for the complete shape and break surface after sampling occupancy values from the fractured shape. Our second contribution is to augment the inference loss for latent code estimation with two penalty terms\textemdash{}(i)~a non-empty restoration term that penalizes the mean restoration occupancy against being zero to avoid empty restorations, and (ii) a proximity term that encourages the mean distance between the complete and restoration occupancy to be low to prevent voluminous restorations. Our work naturally yields the restoration occupancy as the conjunction of the complete occupancy and the negation of the break occupancy, enabling its reconstruction using Marching Cubes~\cite{lorensen1987marching}.

We train and test our approach on synthetically fractured meshes from 8 classes from the ShapeNet~\cite{shapenet2015} dataset, and on the Google Scanned Objects dataset~\cite{google2021scanned} which contains 1,032 scanned real-world objects. We use ShapeNet-trained networks to restore synthetically fractured meshes from the QP Cultural Heritage dataset~\cite{koutsoudis2009qp}, and to generate restorations for physically fractured and scanned real-world mugs. We compare our work to 3D-ORGAN~\cite{hermoza20183d}, the only existing automated fracture restoration approach, and to two baselines. We show state-of-the-art results in overall accuracy and avoiding inaccurate artifacts over non-fracture regions. Our code is available at \url{https://github.com/Terascale-All-sensing-Research-Studio/DeepMend}.

\section{Related Work}

\textbf{Restoration of Fractured Shapes.} Most existing approaches to generate restoration shapes from fractured shapes rely on shape symmetry~\cite{papaioannou2017reassembly, gregor2014towards}. They restore shapes by reflecting non-fractured regions of the shape onto fractured regions and computing the subtraction. These approaches fail to restore asymmetrical shapes or shapes that have non-symmetric fractures. Lamb et al.~\cite{lamb2019automated} perform repair without relying on symmetries. However, they require that the complete counterpart be provided as input alongside the fractured shape. The complete shape may not always be available, e.g., in the case of a rare object. Our work only requires the fractured shape as input. 3D-ORGAN~\cite{hermoza20183d} performs shape restoration in voxel space by using a voxelized representation of an input fractured shape as input to a generative adversarial network. 3D-ORGAN operates at a resolution of 32$\times$32$\times$32, which is insufficient to accurately represent the geometric complexity of the fracture region. Scaling 3D-ORGAN to a voxel resolution necessary to represent fracture is impractical at current dataset volumes and hardware. DeepMend overcomes the challenges of 3D-ORGAN by using networks that represent point samples of the occupancy function. 

\textbf{Completion of Partial Shapes.} Though not directly related to our work, a large body of prior work focuses on completing shapes from partial shape representations, e.g. depth maps or color images. Recent approaches hypothesize complete shapes from partial shapes using deep networks. Approaches that use point clouds as input~\cite{dai2017shape,han2017high,achlioptas2018learning, yuan2018pcn, sarmad2019rl, liu2020morphing, son2020saum, pan2021variational} lack an intrinsic surface representation. Some approaches predict 3D meshes~\cite{yu2022part, groueix2018papier} to incorporate surfaces. These approaches are limited in the complexity of meshes reliably predicted~\cite{mescheder2019occupancy}, and cannot represent arbitrary topologies. Most approaches using voxels~\cite{brock2016generative, sharma2016vconv, wu2016learning, smith2017improved} struggle to predict high-resolution outputs while being computationally tractable. Some voxel approaches address computational inefficiency by employing hierarchical models~\cite{dai2020sg, dai2018scancomplete} or sparse convolutions~\cite{dai2020sg, yi2021complete}. However, voxel approaches pre-discretize the domain, making it challenging to use them to represent arbitrarily fine resolutions needed for geometric detail, especially for the problem of fracture surface representation considered in this work. 

A large body of recent work focuses on using neural networks to represent point samples of values that implicitly define surfaces, e.g., occupancy~\cite{mescheder2019occupancy, chen2019learning, peng2020convolutional, jia2020learning, lionar2021dynamic, liao2018deep, yan2022shapeformer, yan2022implicit, sulzer2022deep, genova2020local, poursaeed2020coupling, chibane2020implicit}, signed distance field (SDF)~\cite{xu2020ladybird, park2019deepsdf, sitzmann2020metasdf, yang2021deep, tretschk2020patchnets, zheng2021deep, lin2020sdf, ma2020neural, xiao2022taylorimnet, duggal2022mending, hao2020dualsdf, chabra2020deep}, unsigned distance field~\cite{tang2021sign}, or level sets~\cite{genova2019learning}. These approaches show high reconstruction fidelity due to their ability to represent the continuous domain of points, while remaining computationally tractable. In contrast to traditional encoder-decoder architectures, approaches based on the autodecoder introduced by DeepSDF~\cite{park2019deepsdf} use maximum \textit{a posteriori} estimation to obtain a latent code for a give shape. The approach enables reconstruction of a complete shape using a latent code estimated using incomplete shape observations. Later approaches provide improvements to the autodecoder by using meta-learning and post-training optimization~\cite{sitzmann2020metasdf, yang2021deep}, and by learning increasingly complex shape representations during training~\cite{duan2020curriculum}, deformation of implicit shape templates~\cite{zheng2021deep}, and reconstruction of shapes at multiple resolutions~\cite{hao2020dualsdf}. 

A potential approach for fractured shape restoration is to convert the fractured shape into a partial shape by removing the fracture surface, perform shape completion, and subtract the fractured shape from the completed shape to obtain the restoration. We demonstrate in Section~\ref{sec:results} that subtraction approaches yield surface artifacts that extend over the non-fracture regions of the fractured shape. Our approach mitigates non-fracture artifacts by learning the interplay between the complete shape and break surface.
\section{Representing Fractured Shapes}

We represent the complete, fractured, and restoration shapes as point sets $C$, $F$, and $R$. For $\mathcal{S} \in \{ C, F, R \}$ the occupancy $o_{\mathcal{S}}(\mathbf{x}) \in \{0, 1\}$ of a point $\mathbf{x}$ is 1 if $\mathbf{x}$ is inside the shape, and 0 if it is on the boundary or outside the shape. The original shapes are closed surfaces. However, we exclude boundary points from the definitions of the sets $C$, $F$, and $R$ to ensure that a point does not simultaneously belong to two sets, e.g., $F$ and $R$. Exclusion of boundary points makes the sets $C$, $F$, and $R$ open and bounded. We define the break surface as a 2D surface that intersects the fracture region of $F$. Points on the side of the break surface corresponding to the fractured shape receive an occupancy of 1. Points on the side corresponding to the restoration shape have an occupancy of 0. We use the open unbounded set $B$, termed the `break set' to represent the set of points with an occupancy $o_B(\mathbf{x})$ of 1. In principle, the break surface is infinite. In practice, we limit the region containing the shape and break sets to be a cube of finite length to make point sampling for network training and inference tractable\footnote{Hereafter, we drop `set' from  references to $C$, $F$, and $R$, and refer to them as shapes.}.

\begin{figure}[t]
    \centering
      \includegraphics[width=.8\linewidth]{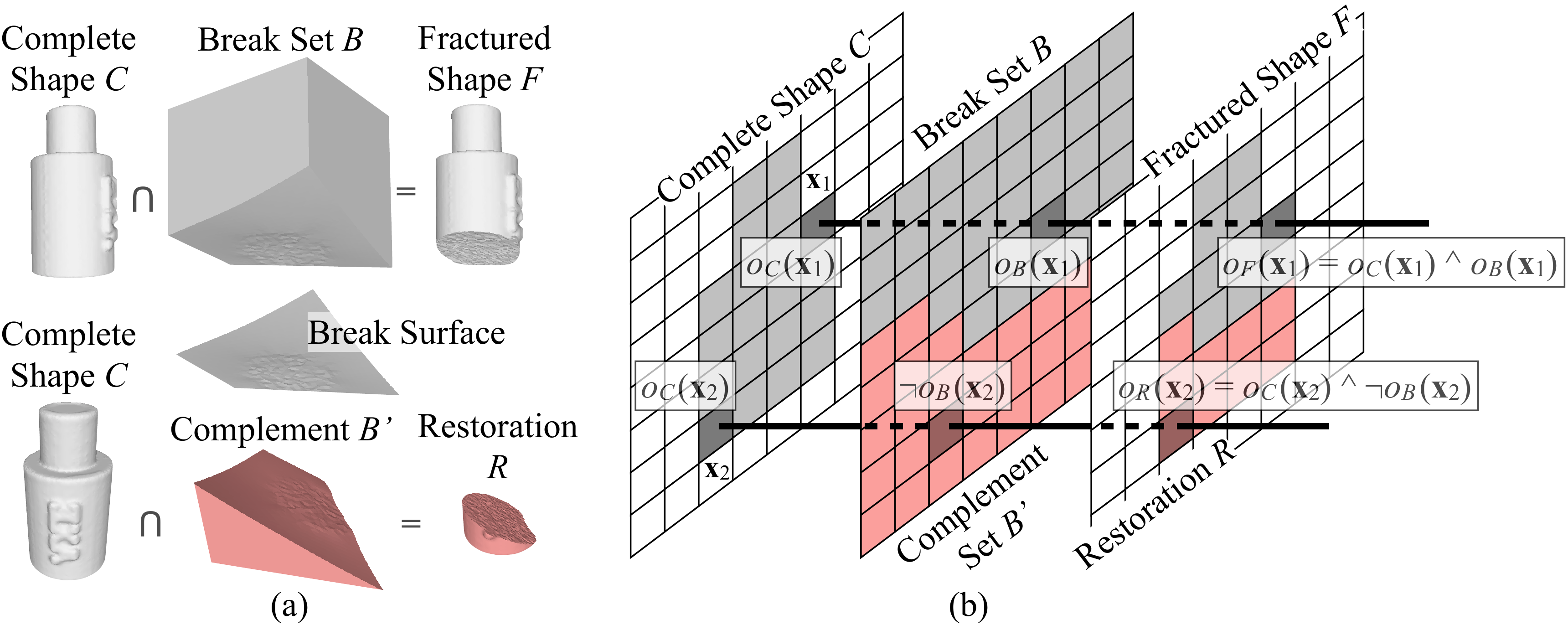}
      \caption{(a) We represent the fractured and restoration shape sets $F$ and $R$ as intersections of the complete shape set $C$ with the break set $B$ and its complement $B^\prime$. (b)~Logical expressions for occupancy at points $\mathbf{x}_1$ in $F$ and $\mathbf{x}_2$ in $R$ expressed in terms of occupancy for $C$ and $B$.}
    \label{fig:reps}
\end{figure}

As shown in Figure~\ref{fig:reps}(a), we represent the fractured shape set as the intersection of the sets for the complete shape and the break set, i.e, as $F = C \cap B$. The set relationship implies that for a point $\mathbf{x}$, occupancy $o_{F}(\mathbf{x})$ is the logical conjunction of the occupancy values $o_{C}(\mathbf{x})$ and $o_{B}(\mathbf{x})$ of the complete shape $C$ and break set $B$, i.e., $o_{F}(\mathbf{x}) = o_{C}(\mathbf{x}) \land o_{B}(\mathbf{x})$. We represent the restoration shape as the intersection of the complete shape and the complement of the break set, i.e, as $R = C \cap B^\prime$. The relationship implies that occupancy $o_{R}(\mathbf{x})$ of the restoration $R$ is expressed as the logical conjunction of $o_{C}(\mathbf{x})$ with the negation of $o_{B}(\mathbf{x})$, i.e., $o_{R}(\mathbf{x}) = o_{C}(\mathbf{x}) \land \neg o_{B}(\mathbf{x})$. The logical relationships are shown in Figure~\ref{fig:reps}(b). To use the expressions in neural networks, we relax the logical relationships using the product T-norm~\cite{gupta1991theory}, as
\begin{align}
    o_{F}(\mathbf{x}) &= o_{C}(\mathbf{x}) o_{B}(\mathbf{x})~\textrm{and}\\
    o_{R}(\mathbf{x}) &= o_{C}(\mathbf{x}) (1 - o_{B}(\mathbf{x})).
    \label{eq:prodr}
\end{align}
We represent the occupancy functions for the complete shape $C$ and break set $B$ respectively using neural networks $f_{\boldsymbol{\Theta}}$ and $g_{\boldsymbol{\Phi}}$, such that $o_{C}(\mathbf{x}) = f_{\boldsymbol{\Theta}}(\mathbf{z}_C, \mathbf{x})$ and $o_{B}(\mathbf{x}) = g_{\boldsymbol{\Phi}}(\mathbf{z}_B, \mathbf{x})$. $\boldsymbol{\Theta}$ and $\boldsymbol{\Phi}$ are the network weights, $\mathbf{z}_C \in \mathbb{R}^p$ is a latent code of size $p$ corresponding to the complete shape, and $\mathbf{z}_B \in \mathbb{R}^q$ is a latent code of size $q$ corresponding to the break surface. We use the autodecoder architecture introduced by Park et al.~\cite{park2019deepsdf} for $f_{\boldsymbol{\Theta}}$ and $g_{\boldsymbol{\Phi}}$. Figure~\ref{fig:inf}(a) shows our network structure. We provide network details in the supplementary.

\subsection{Network Training}
During training, we use a dataset containing multiple training samples to optimize for network parameters $\boldsymbol{\Theta}$ and $\boldsymbol{\Phi}$, and the latent codes $\mathbf{z}_B$ and $\mathbf{z}_C$ for each training sample. Each sample consists of a tuple $(F,C,R,B)$ representing the fractured, complete, and restoration shapes $F$, $C$, and $R$, and the break set $B$ for the sample. We define the training loss as
\begin{align}
    \mathcal{L} &= \textstyle\sum_{\mathbf{z}_{C} \in {Z}_{C}, \mathbf{z}_{B} \in {Z}_{B}} \mathcal{L}_{F} + \mathcal{L}_{C}  + \mathcal{L}_{R} + \mathcal{L}_{B} + \lambda_{\textrm{reg}}\mathcal{L}_{\textrm{reg}}
    \label{eq:trloss}
\end{align}
where ${Z}_{C}$ is the set of all training complete latent codes, and ${Z}_{B}$ is the set of all training break latent codes. The term $\mathcal{L}_{F}$, represented as
\begin{align}
    \mathcal{L}_{F} &= (1/|X|) \textstyle\sum_{\mathbf{x} \in X} {BCE}\left( f_{\boldsymbol{\Theta}}(\mathbf{z}_C, \mathbf{x}) g_{\boldsymbol{\Phi}}(\mathbf{z}_B, \mathbf{x}), o_{F}(\mathbf{x}) \right),
    \label{eq:trfloss}
\end{align}
models the reconstruction of the fracture shape occupancy values. $BCE$ represents the binary cross-entropy loss function. The first argument to ${BCE}$ represents the occupancy expression from Equation~(2), with the expressions for the complete and break occupancy values in terms of $f_{\boldsymbol{\Theta}}$ and $g_{\boldsymbol{\Phi}}$ substituted in. The second argument represents fractured shape ground truth occupancy values. $X$ represents the set of point samples used to probe the ground truth occupancy values. We include terms $\mathcal{L}_C$, $\mathcal{L}_B$, and $\mathcal{L}_R$ to improve the  representation capability of the network by using ground truth occupancy values from training complete shapes, break surfaces, and restorations. We define $\mathcal{L}_{C}$ and $\mathcal{L}_B$ as
\begin{align}
    \mathcal{L}_{C} &= (1/|X|) \textstyle\sum_{\mathbf{x} \in X} 
        {BCE}\left( f_{\boldsymbol{\Theta}}(\mathbf{z}_C, \mathbf{x}), o_{C}(\mathbf{x}) \right)~\textrm{and}\\
    \mathcal{L}_{B} &= (1/|X|) \textstyle\sum_{\mathbf{x} \in X} 
        {BCE}\left( f_{\boldsymbol{\Theta}}(\mathbf{z}_B, \mathbf{x}), o_{B}(\mathbf{x}) \right)
    \label{eq:5}
\end{align}
In Equations (5) and (6), the first argument to $BCE$ represents the occupancy for the complete shape and break set respectively expressed in terms of $f_{\boldsymbol{\Theta}}$ and $g_{\boldsymbol{\Phi}}$. The second argument represents ground truth occupancy values for the complete shape $C$ and break set $B$. We define $\mathcal{L}_{R}$ as 
\begin{align}
    \mathcal{L}_{R} &= (1/|X|) \textstyle\sum_{\mathbf{x} \in X} {BCE}\left( f_{\boldsymbol{\Theta}}(\mathbf{z}_C, \mathbf{x}) (1-g_{\boldsymbol{\Phi}}(\mathbf{z}_B, \mathbf{x})), o_{R}(\mathbf{x}) \right).
    \label{eq:trrloss}
\end{align}
The first argument to ${BCE}$ in Equation~\eqref{eq:trrloss} represents the occupancy expression for restoration shapes from Equation~\eqref{eq:prodr}, with expressions for complete and break occupancy in terms of $f_{\boldsymbol{\Theta}}$ and $g_{\boldsymbol{\Phi}}$ substituted in. The second argument represents ground truth restoration occupancy values. $\mathcal{L}_{\textrm{reg}}= \left\| \mathbf{z}_{C} \right\|_1 + \left\| \mathbf{z}_{B} \right\|_1$ regularizes latent code estimation by imposing a zero-mean Laplacian prior on the latent codes. We set the weight $\lambda_{\textrm{reg}}$ on $\mathcal{L}_{\textrm{reg}}$ to be $1e-4$.

\subsection{Inference of Latent Codes}

\begin{figure}[t]
    \centering
      \includegraphics[width=\linewidth]{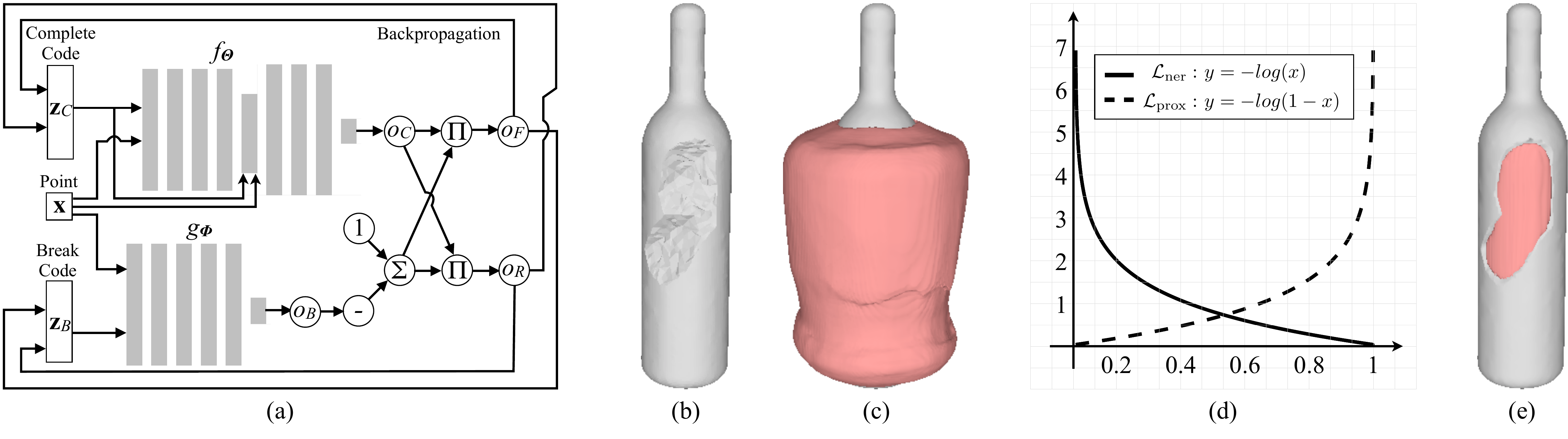}
      \caption{(a) Networks $f_{\boldsymbol{\Theta}}$ and $g_{\boldsymbol{\Phi}}$ represent the complete and break occupancy in terms of input point $\mathbf{x}$ and latent codes $\mathbf{z}_{C}$ and $\mathbf{z}_{B}$. The fractured and restoration occupancy values are composed using the product T-norm relaxations of logical operations for occupancy. Backpropagation updates network weights and latent codes during training. (b) Predicted restoration shape with high non-empty penalty (c) and with high proximity penalty. (d) Functions used for the non-empty penalty term $\mathcal{L}_{\textrm{ner}}$ and the proximity penalty term $\mathcal{L}_{\textrm{prox}}$ during inference. (e) Predicted restoration with balanced penalties. Restoration shapes (red) shown with ground truth fractured shapes (gray).}
    \label{fig:inf}
\end{figure}

During inference, we estimate optimal latent codes $\mathbf{z}_C$ and $\mathbf{z}_B$ for point observations of occupancy $o_{F}(\mathbf{x})$ from a novel input fractured shape $F$. With knowledge of the fractured shape occupancy, the inference loss is given as $\mathcal{L}_\textrm{inf} = \mathcal{L}_F + \lambda_\textrm{reg} \mathcal{L}_\textrm{reg}$. By itself, the loss does not prevent the break surface from being predicted outside or on the boundary of the complete shape. This may result in an empty restoration shape as shown in Figure~\ref{fig:inf}(b). The loss also does not constrain the restoration shape from growing arbitrarily large. Gradient descent on the loss may generate a locally optimal latent code that yields a plausible complete shape, but a large restoration as shown in Figure~\ref{fig:inf}(d). We introduce two penalty terms that encourage  point occupancy values for the restoration that constrain its structure. The non-empty penalty term $\mathcal{L}_\textrm{ner}$, given as
\begin{align}
    \mathcal{L}_{\textrm{ner}} = - \log \left( (1/|X|) \textstyle\sum_{\mathbf{x} \in X} f_{\boldsymbol{\Theta}}(\mathbf{z}_C, \mathbf{x}) (1-g_{\boldsymbol{\Phi}}(\mathbf{z}_B, \mathbf{x})) \right),
\end{align}
penalizes the mean restoration occupancy against being zero. The term encourages the complete shape to have a non-empty intersection with the break set on the restoration side of the break surface. The proximity loss, $\mathcal{L}_{\textrm{prox}}$, given as
\begin{align}
    \mathcal{L}_{\textrm{prox}} = - \log \left( 1 - (1/|X|) \textstyle\sum_{\mathbf{x} \in X} ( f_{\boldsymbol{\Theta}}(\mathbf{z}_C, \mathbf{x}) - o_{F}(\mathbf{x}) )^2 \right),
\end{align}
penalizes the network from predicting complete shapes that are not in close proximity to the fractured shape. The term discourages voluminous restorations. As shown in Figure~\ref{fig:inf}(a), the negative log functions for $\mathcal{L}_{\textrm{ner}}$ and    $\mathcal{L}_{\textrm{prox}}$ strongly penalize the mean occupancy from being too low or the mean complete-restoration occupancy distance from being too high. Using the non-empty and proximity penalties, we express the augmented inference loss $\mathcal{L}_\textrm{infaug}$ as
\begin{equation}
    \mathcal{L}_\textrm{infaug} = \mathcal{L}_\textrm{inf} + \lambda_\textrm{ner} \mathcal{L}_\textrm{ner} + \lambda_\textrm{prox} \mathcal{L}_\textrm{prox} + \lambda_\textrm{reg} \mathcal{L}_\textrm{reg}
    \label{eq:infaug}
\end{equation} 
where $\lambda_{\textrm{ner}}$ and  $\lambda_{\textrm{prox}}$ are weights on the non-empty and proximity penalties. During code estimation, we use $\lambda_{\textrm{ner}} = 1e-5$, $\lambda_{\textrm{prox}} = 5e-3$, and $\lambda_{\textrm{reg}} = 1e-4$. We optimize $\mathcal{L}_\textrm{infaug}$ to estimate the complete and break codes $\mathbf{z}_C$ and $\mathbf{z}_B$. We use the estimated codes to reconstruct restoration occupancy values using Equation~\eqref{eq:prodr}, and obtain the restoration shape as a 3D mesh using Marching Cubes~\cite{lorensen1987marching}.

\section{Datasets and Data Preparation}
\label{sec:dataprep}

We evaluate our work using 3D object models from four datasets.
\begin{enumerate}
    \item \textbf{ShapeNet.} We use 3D meshes from 8 classes in the ShapeNet dataset~\cite{shapenet2015} of synthetic 3D models: airplanes, bottles, cars, chairs, jars, mugs, sofas, and tables. Each class has between 1,345 to 5,324 shapes, with an average of 3,084 shapes per class. We create one network per class, and use an 80\%/20\% train/test split of the meshes within each class.
    \item \textbf{Google Scanned Objects Dataset.} The dataset~\cite{google2021scanned} contains 1,032 digitally scanned common objects such as cups, bowls, plates, baskets, and shoes. We train a network with an 80\%/20\% train/test split of the dataset.
    \item \textbf{QP Cultural Heritage Object Dataset.} The QP dataset~\cite{koutsoudis2009qp} contains 317 meshes computer-modeled in the style of ancient Greek pottery. We use all models for testing using the network trained on ShapeNet jars.
    \item \textbf{Real-World Fractured Mugs.} We perform in-house fractures of 4 real-world mugs, and scan the fractured mugs for testing. We use all mug models for testing using the network trained on ShapeNet mugs.
\end{enumerate}

\begin{figure}[t]
    \centering
      \includegraphics[width=\linewidth]{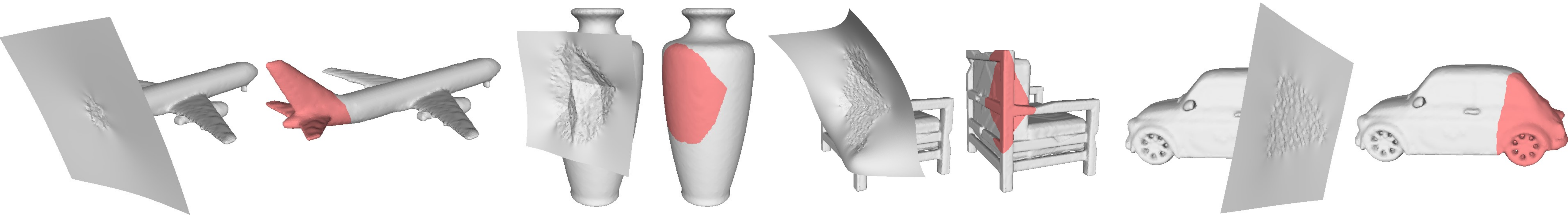}
      \caption{Ground truth fractured shapes shown with fitted thin-plate splines (TPS) corresponding to the surface of the break surface, and with ground truth restoration shapes. Fractured shapes and TPS are shown in gray, restoration shapes are shown in red.}
    \label{fig:dataprep}
\end{figure}

\textbf{Data Preparation.} We center meshes and scale them to lie within a unit cube. ShapeNet and QP models come pre-oriented to be consistent. Though meshes for Google Scanned Objects have a common ground plane, they are not oriented in a consistent direction. We augment the training set for Google Scanned Objects by randomly rotating meshes by 90$^\circ$ around the ground plane normal. We orient all real-world mugs to line up with ShapeNet mugs. We waterproof meshes using the approach of Stutz and Geiger~\cite{Stutz2018ARXIV}. 

The ShapeNet, Google Scanned Objects, and QP Cultural Heritage datasets lack fractures. We synthetically fracture meshes in these datasets by repeatedly subtracting a randomized geometric primitive from each mesh. We use the fracture approach from Lamb et al.~\cite{lamb2021using}, except that we fracture each mesh such that between 5\% and 20\% of the surface area of the mesh is removed by the fracture. We show example fractured shapes, restorations, and break surfaces in Figure~\ref{fig:dataprep}. We remove shapes from the test set with more than one connected component. The mugs, jars, and bottles classes from ShapeNet have fewer than 600 meshes. We fracture meshes belonging to these classes 10, 3, and 3 times respectively. We augment the mugs set by requiring that 3 fractures for each complete mesh only remove parts of the handle. We fracture meshes in the remaining ShapeNet classes and the QP and Google Scanned Objects datasets once.

We obtain point samples for the set $X$ by randomly sampling a unit cube around the object and the surface of the mesh as described in the supplementary. To generate the break surface for each training sample, we fit a thin-plate spline (TPS)~\cite{duchon1977splines} to the fracture vertices, such that the spline domain corresponds to the closest fitting plane to the fracture region vertices. We use the spline to partition sample points in the interior of the fractured and restoration meshes into two groups. We denote the side of the spline that contains the most fractured shape sample points as belonging to the break set $B$.

\section{Results}
\label{sec:results}

\textbf{Metrics.} We perform evaluations using the chamfer distance (CD), percentage of non-empty restorations (NE\%), and non-fracture region error (NFRE). NFRE measures the degree to which the restoration surface incorrectly contains geometry close to the non-fracture region of the fractured shape. To compute NFRE, we sample $n$ points on the surfaces of the predicted restoration, ground truth restoration, and non-fractured region of the fractured shape, the predicted restoration. We count points on the predicted restoration that have a nearest neighbor in the non-fracture region of the fractured shape closer than $\eta$ and a nearest neighbor in the ground truth restoration farther than $\eta$. We normalize the count by $n$. We use $\eta=0.02$ and $n=30,000$. For all metrics, we report means over all classes as weighted to overcome class imbalances.

\subsection{Results using ShapeNet}

\begin{figure}[t]
    \centering
      \includegraphics[width=\linewidth]{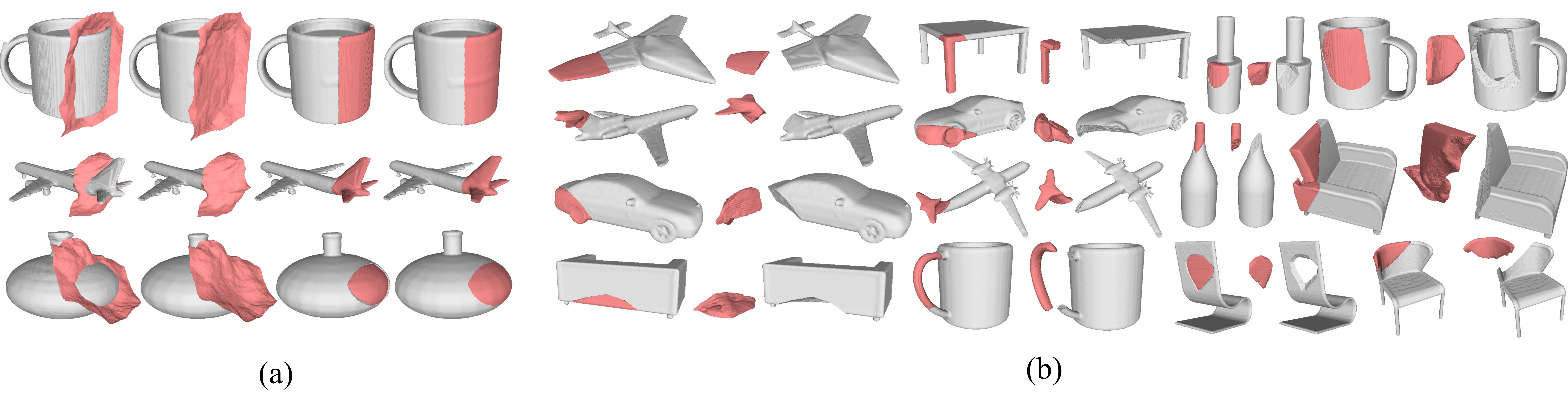}
      \caption{(a) Predicted break surface and predicted complete shape, predicted break surface and fractured shape, predicted restoration shape and fractured shape, and ground truth fractured and restoration shape for three objects. (b) Predicted restoration shapes, shown with ground truth fractured shapes in joined form and opened to show the fracture surface. Fractured shapes are in gray, restoration shapes in red.}
    \label{fig:ours}
\end{figure}

Figure~\ref{fig:ours}(a) shows DeepMend-generated complete shapes, and predicted break surfaces and restoration shapes joined to input fractured shapes synthesized from ShapeNet objects models. The break surfaces predicted by DeepMend  mimic the fracture region at the join, resulting in accurate connections between the input fractured shape and the predicted restoration shape. Figure~\ref{fig:ours}(b) shows restorations joined to corresponding ground truth fractured shapes and opened to show the fracture region. DeepMend restoration shapes match closely to the fractured shape, and avoid surface artifacts that may otherwise prevent the restoration shape from being physically joined to the fractured shape. DeepMend regenerates complex missing geometry, such as the tails of the planes in Figure~\ref{fig:ours}(b) and the car spoiler in Figure~\ref{fig:teaser}. It restores complex fractures such as the hole in the back of the lawn chair in Figure~\ref{fig:ours}(b). In contrast to symmetry approaches~\cite{papaioannou2017reassembly, gregor2014towards}, DeepMend repairs asymmetrical shapes or shapes with asymmetrical fractures such as the sofa, planes with broken tails on left, bottle, armchair, lawn chair, and plastic chair in Figure~\ref{fig:ours}, and the car, mug, sofa, and airplane in Figure~\ref{fig:teaser}.

\textbf{Effect of Penalties on Restoration Shape During Inference.} We evaluate the impact of our augmented inference loss $\mathcal{L}_\textrm{infaug}$ in Equation~\eqref{eq:infaug} versus adding no penalty terms, adding solely the non-empty term $\mathcal{L}_{\textrm{ner}}$, and adding solely the proximity term $\mathcal{L}_{\textrm{prox}}$. We also evaluate alternative penalties given as
\begin{align}
\mathcal{L}_{\textrm{nerp}} &= - (1/|X|) \textstyle\sum_{\mathbf{x} \in X}\log \left(  f_{\boldsymbol{\Theta}}(\mathbf{z}_C, \mathbf{x}) (1-g_{\boldsymbol{\Phi}}(\mathbf{z}_B, \mathbf{x})) \right)~\textrm{and}\\
\mathcal{L}_{\textrm{proxp}} &=  (1/|X|) \textstyle\sum_{\mathbf{x} \in X}BCE(f_{\boldsymbol{\Theta}}(\mathbf{z}_C,\mathbf{x}),o_{F}(\mathbf{x})),
\end{align}
that penalize individual points rather than mean occupancy or distance. $\mathcal{L}_{\textrm{nerp}}$ discourages individual points from being 0. $\mathcal{L}_{\textrm{proxp}}$ encourages individual occupancy values in $C$ to be similar to $F$. 

\begin{figure}[t]
    \centering
      \includegraphics[width=\linewidth]{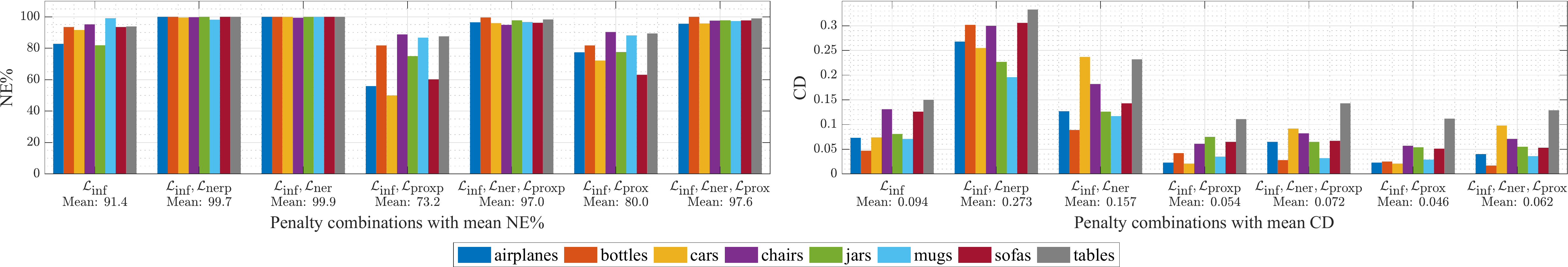}
      \caption{Percentage of non-empty restorations (NE\%, left) and chamfer distance (CD, right) for our approach with various combinations of penalties for the inference loss.}
    \label{fig:infloss}
\end{figure}

\begin{figure}[t]
    \centering
      \includegraphics[width=.8\linewidth]{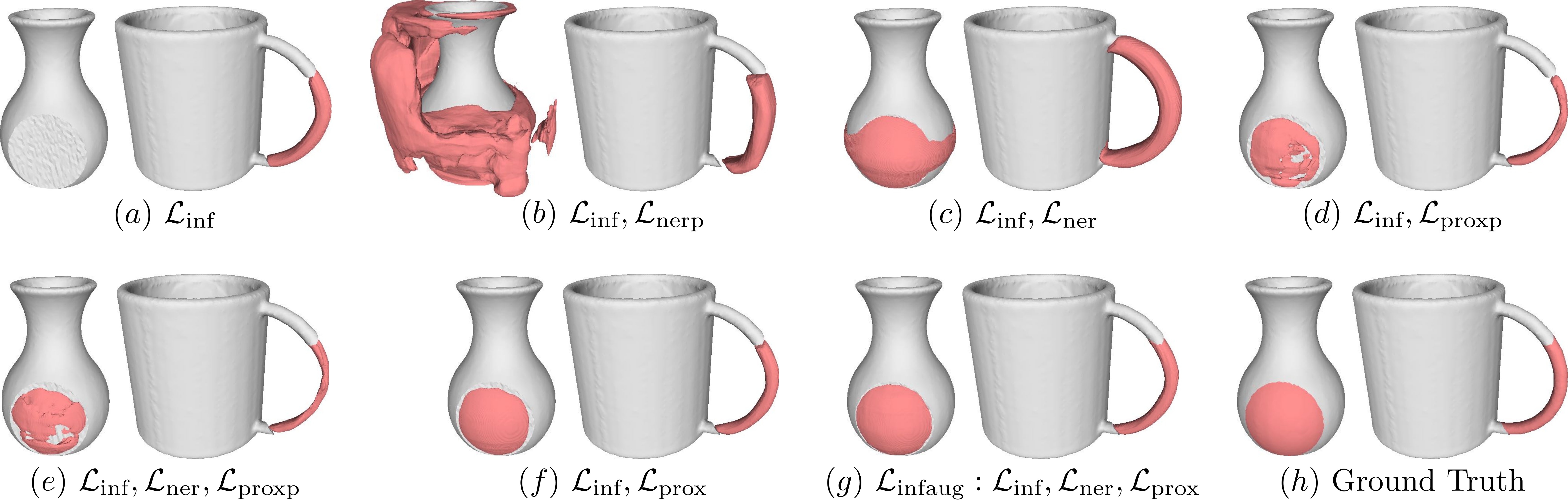}
      \caption{Predicted restoration shapes (red) shown with ground truth fractured shapes using (gray) using various combinations of penalty terms.}
    \label{fig:loss}
\end{figure}

We summarize results in the bar plots in Figure~\ref{fig:infloss}. Using no penalties predicts large restoration shapes, or generates multiple empty restorations as shown by the jar in Figure~\ref{fig:loss}(a). Mean NE\% and CD with $\mathcal{L}_\textrm{inf}$ are 91.5\% and 0.096. Including $\mathcal{L}_\textrm{nerp}$ raises mean NE\% to 99.7\%. However, since the penalty is applied to individual points, restorations appear splayed out and non-smooth as shown by the jar in Figure~\ref{fig:loss}(b). Mean CD is the highest with $\mathcal{L}_\textrm{nerp}$ at 0.275. By penalizing mean occupancy, $\mathcal{L}_\textrm{ner}$ remedies the splaying by keeping occupancy values concentrated, improves restoration quality, and lowers CD to a mean of 0.159. However, restorations may now appear bulkier as shown by the mug in Figure~\ref{fig:loss}(c). Mean CD is higher than when no penalty term is used. 

When comparing the effect of the proximity penalties on inference, we find that including $\mathcal{L}_{\textrm{proxp}}$ improves mean CD to 0.055. However, it considerably drops the percentage of non-empty restorations to a mean of 73.4\%. The per-point penalty induces individual points to approach the fracture surface, causing restorations to lose volume and become non-smooth  as shown in Figure~\ref{fig:loss}(d). Inclusion of the non-empty restoration term $\mathcal{L}_{\textrm{ner}}$ improves NE\% to 97.0\% with minimal impact on CD. However, it fails to remedy the non-smooth shape as shown by the jar in Figure~\ref{fig:loss}(e). Several individual points still show zero occupancy values. By using the penalty on the mean complete-restoration proximity $\mathcal{L}_{\textrm{prox}}$, per-point distances remain concentrated and NE\% is higher than with $\mathcal{L}_{\textrm{proxp}}$. Combining both proposed penalties improves NE\% over solely using $\mathcal{L}_{\textrm{prox}}$ from 80.1\% to 97.6\%, without compromising on the mean CD at 0.064. As shown by Figure~\ref{fig:loss}(g), using penalties on the mean occupancy and proximity values provides balanced, concentrated, and smooth restorations. 

\subsection{Comparing DeepMend Restorations to 3D-ORGAN \& Baselines}

We compare DeepMend restorations using ShapeNet to 3D-ORGAN \cite{hermoza20183d}, the only publicly available approach to restore fractured inputs. We also compare to two baseline approaches.

\textbf{3D-ORGAN.} 3D-ORGAN~\cite{hermoza20183d} takes a voxelized fractured shape at a resolution of $32\times32\times32$ as input and predicts the corresponding complete voxelized shape. We adapt 3D-ORGAN to generate restoration shapes by computing the restoration shape as the element-wise difference between the complete voxels predicted by 3D-ORGAN and the input fractured voxels. We train 3D-ORGAN on fractured shapes from 8 classes from ShapeNet. During training, we compute the fractured shapes randomly by removing voxel regions from each shape according to the original implementation. During testing, we predict complete shapes using voxelized ShapeNet fractured shapes generated as described in Section~\ref{sec:dataprep}. As recommended by the authors, we use a two-step approach to predict complete shapes by feeding the complete shape predicted in the first iteration as input to 3D-ORGAN for a second iteration.

\setlength{\tabcolsep}{2.1pt}
\begin{table*}[t!]
\scriptsize
\centering
\begin{tabular}{c|c|cccccccc|c}
\toprule
 & Metric & airplanes & bottles & cars & chairs & jars & mugs & sofas & tables & Mean \\ \hline \hline
\multirow{2}{*}{3D-ORGAN} & CD & 0.194 & 0.146 & - & 0.195 & 0.282 & - & 0.329 & 0.352 & 0.250 \\
 & NFRE & 0.192 & 0.066 & - & 0.585 & 0.036 & - & 0.199 & 0.137 & 0.203 \\ \hline
\multirow{2}{*}{Sub-Occ} & CD & 0.044 & 0.034 & \textbf{0.019} & 0.096 & 0.128 & 0.037 & 0.062 & 0.118 & 0.067 \\
 & NFRE & 0.089 & 0.058 & 0.115 & 0.239 & 0.133 & 0.055 & 0.160 & 0.175 & 0.128 \\ \hline
\multirow{2}{*}{Sub-Lamb} & CD & 0.080 & 0.032 & 0.037 & 0.073 & 0.077 & 0.121 & \textbf{0.044}  & \textbf{0.087} & 0.069 \\
 & NFRE & 0.294 & 0.085 & 0.217 & 0.296 & 0.259 & 0.487 & 0.168 & 0.193 & 0.250 \\  \hline
\multirow{2}{*}{DeepMend (Ours)} & CD & \textbf{0.040} & \textbf{0.017} & 0.098 & \textbf{0.071} & \textbf{0.055} & \textbf{0.036} & 0.053 & 0.130 & \textbf{0.063} \\
 & NFRE & \textbf{0.009} & \textbf{0.010} & \textbf{0.017} & \textbf{0.009} & \textbf{0.007} & \textbf{0.008} & \textbf{0.012} & \textbf{0.012} & \textbf{0.011} \\
 \bottomrule
\end{tabular}
\caption{Chamfer (CD) distance and NFRE when comparing 3D-ORGAN, Sub-Occ, Sub-Lamb, and DeepMend. For 3D-ORGAN, we do not report results for cars and mugs, as it only restores 21.7\% cars and 1.8\% mugs. Overall mean values provided over classes for which we report individual class means. Bold values correspond to the best performing metric value within a class.}
\label{tab:deepsdf}
\end{table*}

\textbf{Baselines of Performing Subtraction from Complete Shape.}
Since no approach exists to automatically generate high-resolution restoration shapes directly from a fractured shape, we adapt partial shape completion to generate a complete shape from which the fractured shape can be subtracted to create a restoration. We generate a partial shape from the fractured shape by removing points detected by a fracture/non-fracture classifier as being fracture points. We train a PointNet++~\cite{qi2017pointnet} classifier to classify points as fracture versus non-fracture. The classifier provides a test accuracy of 85.6\%. We remove detected fracture points to generate the partial shape. We train DeepSDF~\cite{park2019deepsdf} on complete shapes for the 8 ShapeNet classes studied in this work. We use DeepSDF to complete the partial shape using the shape completion method discussed by the authors. We use the following two approaches for subtraction as baselines.
\begin{enumerate}
    \item \textbf{Sub-Occ.} We convert SDF values for the input fractured shape and DeepSDF-predicted complete shape into occupancy values. We take the difference of the complete and fractured occupancy, and extract the 0-level isosurface. To remove artifacts, we discard closed surfaces with a volume less than $\eta=0.01$. If this step removes all closed surfaces, we retain the largest surface. 
    \item \textbf{Sub-Lamb.} We use the approach of Lamb et al.~\cite{lamb2019automated} to perform subtraction. Unlike our approach, Lamb et al. require a complete counterpart to be provided alongside the fractured shape as input. We use the complete shape predicted using DeepSDF as the complete counterpart. We repair self-intersections at the fracture-restoration join using MeshFix~\cite{attene2010lightweight}.
\end{enumerate}

\begin{figure}[t]
    \centering
      \includegraphics[width=\linewidth]{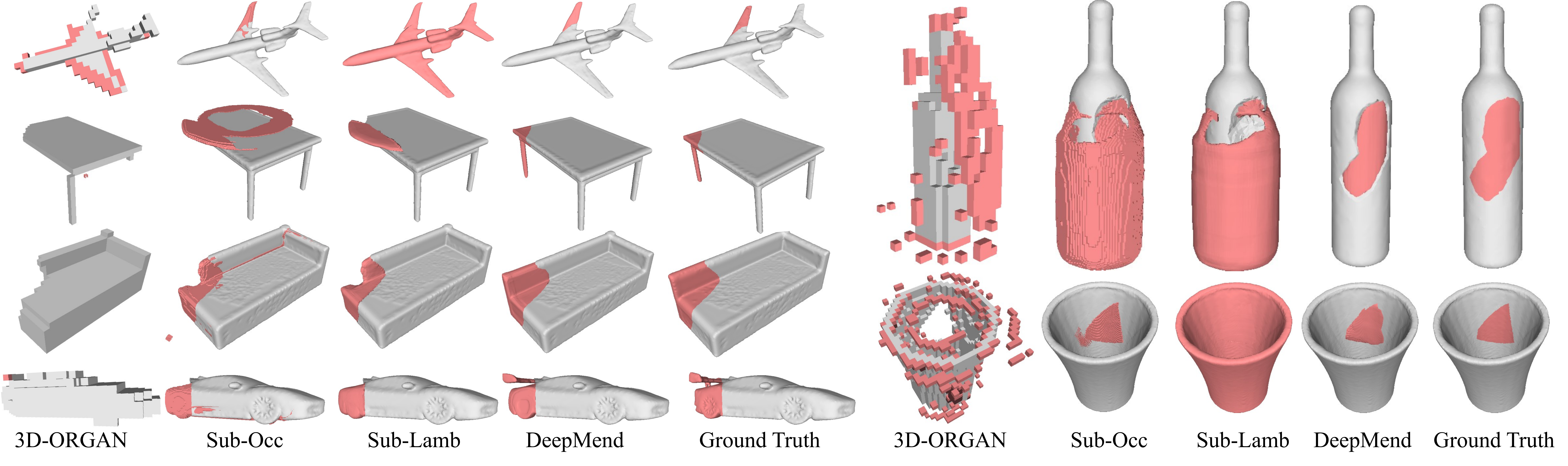}
      \caption{Pictorial results of restorations using 3D-ORGAN, baselines Sub-Occ and Sub-Lamb, and DeepMend, together with ground truth restorations.}
    \label{fig:compare}
\end{figure}

Table~\ref{tab:deepsdf} summarizes results of CD and NFRE using 3D-ORGAN, Sub-Occ, Sub-Lamb, and DeepMend. 100\% restorations are generated by Sub-Occ and Sub-Lamb for all classes, and by 3D-ORGAN for all classes except cars and mugs. We report results for over all shapes where DeepMend returns non-empty restorations for all classes where Sub-Occ, Sub-Lamb, and 3D-ORGAN provide 100\% restorations. For 3D-ORGAN, we exclude metrics for cars and mugs, as 3D-ORGAN only produces restorations for 6 out of 331 or 1.8\% of mug shapes and 103 out of 474 or 21.7\% car shapes. Figure~\ref{fig:compare} shows visualizations of the results produced using 3D-ORGAN, Sub-Occ, Sub-Lamb, and DeepMend.

\begin{figure}[t]
    \centering
      \includegraphics[width=\linewidth]{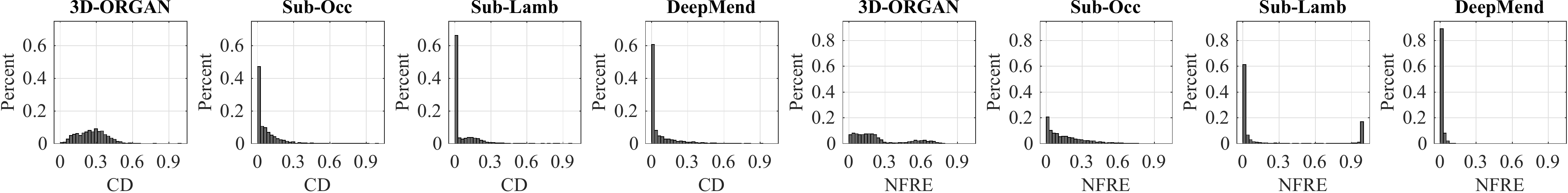}
      \caption{Histograms of chamfer distance (CD, first four plots) and non-fracture region error (NFRE, last four plots) for 3D-ORGAN, Sub-Occ, Sub-Lamb, and Ours.}
    \label{fig:comparehist}
\end{figure}

DeepMend shows state-of-the-art results compared to 3D-ORGAN and the two baselines in terms of overall CD and NFRE, and in terms of per-class mean NFRE. DeepMend also shows lower CD for all classes than 3D-ORGAN, 6 out of 8 classes in comparison to Sub-Occ, and 5 out of 8 classes when compared to Sub-Lamb. 3D-ORGAN predicts small restoration shapes, as shown for the table, sofa, and car in Figure~\ref{fig:compare}, or disparate voxels surrounding the fractured shape, as shown for the airplane, bottle, and jar in Figure~\ref{fig:compare}. The histogram on the left of Figure~\ref{fig:comparehist} shows that 3D-ORGAN predicts 0.5\% restorations with a chamfer distance less than 0.025, compared to 63.1\% with DeepMend.

Restoration shapes generated using Sub-Occ exhibit artifacts on the surface of the fractured shape as shown in Figure~\ref{fig:compare}. The histogram of the NFRE values for Sub-Occ in Figure~\ref{fig:comparehist} shows that 23.3\% of restorations have NFRE lower than 0.025, as opposed to 88.3\% by DeepMend. The fracture classifier may not reliably remove the entire fracture region to create a partial shape that is a precise subset of the complete shape. As such, Sub-Occ generates restorations that exhibit elements of the fracture, e.g., the sofa, car, and bottle in Figure~\ref{fig:compare}, or predicts the restoration in the wrong location, e.g., the table in Figure~\ref{fig:compare}.

As shown in Figure~\ref{fig:comparehist}, Sub-Lamb is effective at removing artifacts for some objects, as 60.8\% of restoration shapes have NFRE lower than 0.025. However, for complex geometries, Sub-Lamb incorrectly marks the exterior region of the fractured shape as belonging to the fracture, causing the entirety of the fractured shape to be merged with the restoration as, e.g., in case of the airplane and jar in Figure~\ref{fig:compare}. The NFRE histogram for Sub-Lamb demonstrates that 18.9\% of restoration shapes have a NFRE between 0.975 and 1. As shown in Figure~\ref{fig:comparehist}, 88.3\% of DeepMend restorations show NFRE lower than 0.025. 

\begin{figure}[t]
    \centering
      \includegraphics[width=\linewidth]{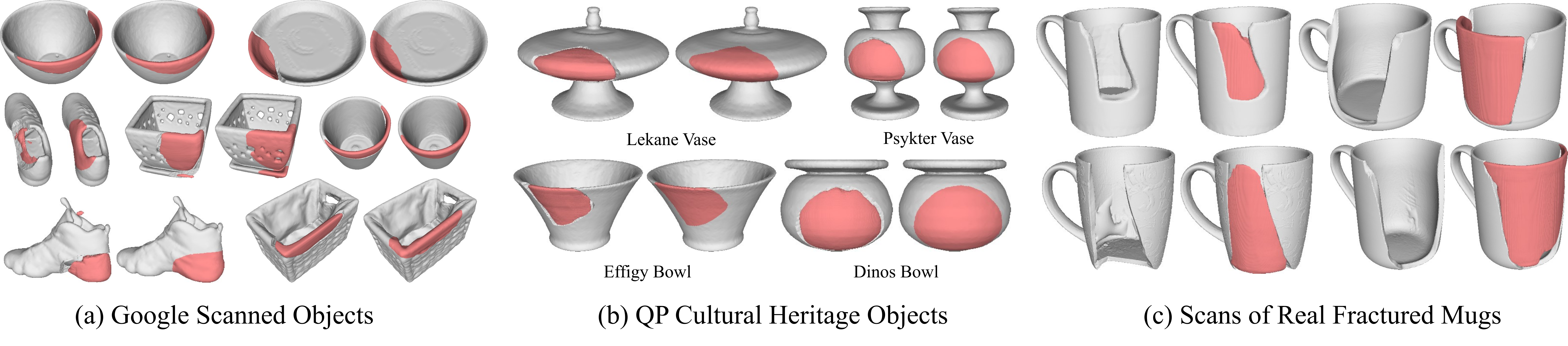}
      \caption{Results using (a) synthetic breaks on Greek pottery from QP~\cite{koutsoudis2009qp}, (b) synthetic fractures on 3D scans from Google Scanned Objects~\cite{google2021scanned}, and (c) 3D scans for real-world fractured mugs. Ground truth restoration shapes shown when applicable.}
    \label{fig:datas}
\end{figure}

\subsection{Results with Google Scanned Objects, QP, and Fractured Mugs}

We show results of training and testing DeepMend with the Google Scanned Objects dataset in Figure~\ref{fig:datas}(a). We obtain a test chamfer distance of 0.112. DeepMend generates closely fitting restorations for objects that are prone to fracture such as the plate, the pot, and the bowls in Figure~\ref{fig:datas}(a), and reasonable restoration shapes for complex shapes with high intra-class variety such as shoes. 

We demonstrate the generalizability of our approach to novel datasets by using ShapeNet-trained jars and mugs networks to restore synthetically fractured shapes for objects from the QP dataset as shown in Figure~\ref{fig:datas}(b), and for 3D scans of 4 real-world fractured mugs as shown in Figure~\ref{fig:datas}(c). We achieve a mean chamfer distance of 0.047 for QP objects. DeepMend generates plausible restorations for shapes that resemble modern bowls, such as the effigy bowl in Figure~\ref{fig:datas}(b), and for uncommon shapes, such as the psykter vase in Figure~\ref{fig:datas}(b).

The restoration process for real-world mugs is particularly challenging. Primitives used to synthesize breaks in the ShapeNet dataset are simplistic. The fractured and non-fracture regions in synthetic breaks have a clear edge. In real-world mugs, the break structure is far more complex with sharp curvature and smoothed out edges as seen in Figure~\ref{fig:datas}(c). Scanning limitations may cause the fracture surface geometry to be less finely captured in comparison to the synthetically-generated roughness of  simulated fractures. Despite the challenges, our approach shows the capability to reconstruct the restoration by generating break surfaces that approach the fracture surface of real broken mugs. 

\section{Conclusion}

\begin{wrapfigure}[11]{r}{0.37\textwidth}
    \centering
    \vspace{-20pt}
      \includegraphics[width=\linewidth]{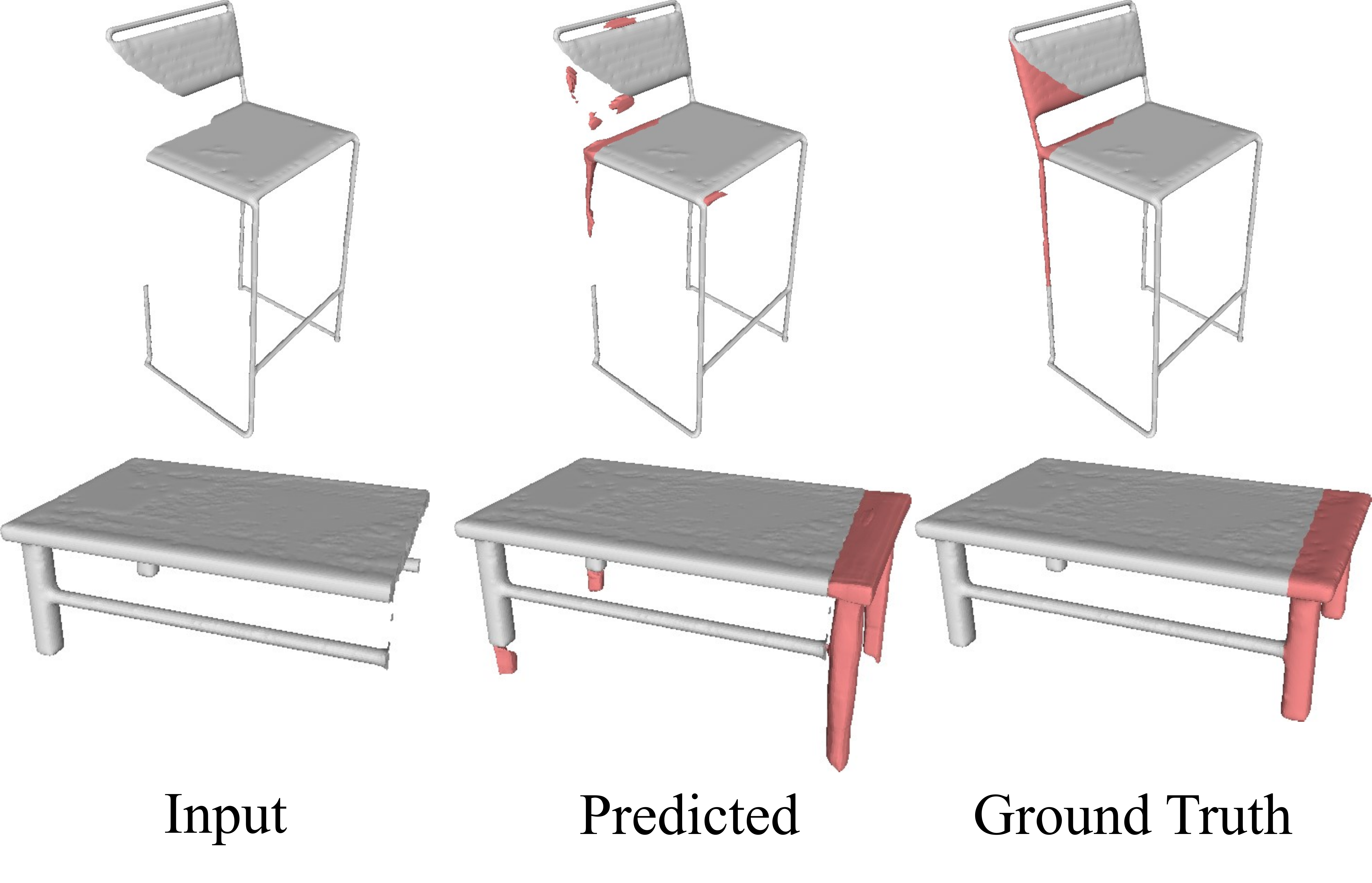}
      \caption{Restorations with multiple components.}
      \vspace{-12pt}
    \label{fig:limit}
\end{wrapfigure}

We provide DeepMend, an approach to automatically restore fractured shapes by learning to represent complete shapes, break surfaces, and their interplay. We contribute penalty functions for inference that penalize mean occupancy values against being too high or low, thereby ensuring  well-structured restorations. DeepMend does not require ground truth knowledge of the fracture region, making it amenable for rapid repair. 

One \textbf{limitation} of our work is that it may predict multiple components for single-component fractures, especially for thin structures, e.g., the chair in Figure~\ref{fig:limit}, which are a common problem for learned functions that represent implicit surface fields. Since the components are on the restoration side of the surface, NFRE remains lower than with the baseline methods. In many cases, e.g. the table in Figure~\ref{fig:limit}, multiple components yield plausible restorations. For the table class, these components together with the high intra-class variance contribute to an increased CD as shown by the results in Table~\ref{tab:deepsdf}. Multiple small component prediction arises as the break surface lacks constraints during inference, and can adopt arbitrarily complex geometries. As part of future work, data-driven priors can be incorporated on the structure of shapes and break surfaces. Future work can use massive datasets to learn prior probability distributions of occupancy of 3D objects. The learned shape representation can be strengthened with intra-object structural constraints, e.g., symmetries and planarity. 

While not explored in this work, the ability of DeepMend to naturally predict multiple components enables it to be leveraged for future work in restoring multiple broken parts of an object. Novel scanned datasets of objects of diverse materials containing varying levels of damage, e.g., chipping, shearing, splintering, and ductile versus brittle fractures can benefit the study of the damage process and impact on fracture surface geometry. Our work contributes a geometric foundation for the study of fractured shape repair using closed 3D surfaces. The work opens the scope for future research on automated shape repair using depth and color images, to facilitate democratization of the repair process.

\clearpage
\bibliographystyle{splncs04}
\bibliography{egbib}

\begin{thebibliography}{10}
\providecommand{\url}[1]{\texttt{#1}}
\providecommand{\urlprefix}{URL }
\providecommand{\doi}[1]{https://doi.org/#1}

\bibitem{achlioptas2018learning}
Achlioptas, P., Diamanti, O., Mitliagkas, I., Guibas, L.: Learning
  representations and generative models for 3d point clouds. In: International
  Conference On Machine Learning. PMLR (2018)

\bibitem{attene2010lightweight}
Attene, M.: A lightweight approach to repairing digitized polygon meshes. The
  visual computer  \textbf{26}(11),  1393--1406 (2010)

\bibitem{brock2016generative}
Brock, A., Lim, T., Ritchie, J.M., Weston, N.: Generative and discriminative
  voxel modeling with convolutional neural networks. arXiv preprint
  arXiv:1608.04236  \textbf{1}(1), ~1--9 (2016)

\bibitem{chabra2020deep}
Chabra, R., Lenssen, J.E., Ilg, E., Schmidt, T., Straub, J., Lovegrove, S.,
  Newcombe, R.: Deep local shapes: Learning local sdf priors for detailed 3d
  reconstruction. In: ECCV. pp. 608--625. Springer, Berlin, Germany (2020)

\bibitem{shapenet2015}
Chang, A.X., Funkhouser, T., Guibas, L., Hanrahan, P., Huang, Q., Li, Z.,
  Savarese, S., Savva, M., Song, S., Su, H., Xiao, J., Yi, L., Yu, F.:
  {ShapeNet: An Information-Rich 3D Model Repository}. Tech. Rep.
  arXiv:1512.03012 [cs.GR], Stanford University --- Princeton University ---
  Toyota Technological Institute at Chicago (2015)

\bibitem{chen2019learning}
Chen, Z., Zhang, H.: Learning implicit fields for generative shape modeling.
  In: Proc. CVPR. pp. 5939--5948. IEEE, Piscataway, NJ (2019)

\bibitem{chibane2020implicit}
Chibane, J., Alldieck, T., Pons-Moll, G.: Implicit functions in feature space
  for 3d shape reconstruction and completion. In: Proc. CVPR. pp. 6970--6981.
  IEEE, Piscataway, NJ (2020)

\bibitem{dai2020sg}
Dai, A., Diller, C., Nie{\ss}ner, M.: Sg-nn: Sparse generative neural networks
  for self-supervised scene completion of rgb-d scans. In: Proceedings of the
  IEEE/CVF Conference on Computer Vision and Pattern Recognition. pp. 849--858
  (2020)

\bibitem{dai2018scancomplete}
Dai, A., Ritchie, D., Bokeloh, M., Reed, S., Sturm, J., Nie{\ss}ner, M.:
  Scancomplete: Large-scale scene completion and semantic segmentation for 3d
  scans. In: Proc. CVPR. pp. 4578--4587. IEEE, Piscataway, NJ (2018)

\bibitem{dai2017shape}
Dai, A., Ruizhongtai~Qi, C., Nie{\ss}ner, M.: Shape completion using
  3d-encoder-predictor cnns and shape synthesis. In: Proceedings of the IEEE
  conference on computer vision and pattern recognition. pp. 5868--5877 (2017)

\bibitem{duan2020curriculum}
Duan, Y., Zhu, H., Wang, H., Yi, L., Nevatia, R., Guibas, L.J.: Curriculum
  deepsdf. In: European Conference on Computer Vision. pp. 51--67. Springer
  (2020)

\bibitem{duchon1977splines}
Duchon, J.: Splines minimizing rotation-invariant semi-norms in sobolev spaces.
  In: Constructive theory of functions of several variables, pp. 85--100.
  Springer (1977)

\bibitem{duggal2022mending}
Duggal, S., Wang, Z., Ma, W.C., Manivasagam, S., Liang, J., Wang, S., Urtasun,
  R.: Mending neural implicit modeling for 3d vehicle reconstruction in the
  wild. In: Proceedings of the IEEE/CVF Winter Conference on Applications of
  Computer Vision. pp. 1900--1909 (2022)

\bibitem{genova2020local}
Genova, K., Cole, F., Sud, A., Sarna, A., Funkhouser, T.: Local deep implicit
  functions for 3d shape. In: Proc. CVPR. pp. 4857--4866. IEEE, Piscataway, NJ
  (2020)

\bibitem{genova2019learning}
Genova, K., Cole, F., Vlasic, D., Sarna, A., Freeman, W.T., Funkhouser, T.:
  Learning shape templates with structured implicit functions. In: Proc. CVPR.
  pp. 7154--7164. IEEE, Piscataway, NJ (2019)

\bibitem{gregor2014towards}
Gregor, R., Sipiran, I., Papaioannou, G., Schreck, T., Andreadis, A., Mavridis,
  P.: Towards automated 3d reconstruction of defective cultural heritage
  objects. In: GCH. pp. 135--144. EUROGRAPHICS, Geneva, Switzerland (2014)

\bibitem{groueix2018papier}
Groueix, T., Fisher, M., Kim, V.G., Russell, B.C., Aubry, M.: A
  papier-m{\^a}ch{\'e} approach to learning 3d surface generation. In: Proc.
  CVPR. pp. 216--224. IEEE, Piscataway, NJ (2018)

\bibitem{gupta1991theory}
Gupta, M.M., Qi, J.: Theory of t-norms and fuzzy inference methods. Fuzzy sets
  and systems  \textbf{40}(3),  431--450 (1991)

\bibitem{han2017high}
Han, X., Li, Z., Huang, H., Kalogerakis, E., Yu, Y.: High-resolution shape
  completion using deep neural networks for global structure and local geometry
  inference. In: Proceedings of the IEEE international conference on computer
  vision. pp. 85--93 (2017)

\bibitem{hao2020dualsdf}
Hao, Z., Averbuch-Elor, H., Snavely, N., Belongie, S.: Dualsdf: Semantic shape
  manipulation using a two-level representation. In: Proceedings of the
  IEEE/CVF Conference on Computer Vision and Pattern Recognition. pp.
  7631--7641 (2020)

\bibitem{hermoza20183d}
Hermoza, R., Sipiran, I.: 3d reconstruction of incomplete archaeological
  objects using a generative adversarial network. In: Proceedings of Computer
  Graphics International, pp. 5--11. ACM, New York, NY (2018)

\bibitem{jia2020learning}
Jia, M., Kyan, M.: Learning occupancy function from point clouds for surface
  reconstruction. arXiv preprint arXiv:2010.11378  \textbf{1}(1),  1--11 (2020)

\bibitem{koutsoudis2009qp}
Koutsoudis, A., Pavlidis, G., Arnaoutoglou, F., Tsiafakis, D., Chamzas, C.: Qp:
  A tool for generating 3d models of ancient greek pottery. Journal of Cultural
  Heritage  \textbf{10}(2),  281--295 (2009)

\bibitem{lamb2019automated}
Lamb, N., Banerjee, S., Banerjee, N.K.: Automated reconstruction of smoothly
  joining 3d printed restorations to fix broken objects. In: Proc. SCF. pp.
  1--12. ACM, New York, NY (2019)

\bibitem{lamb2021using}
Lamb, N., Wiederhold, N., Lamb, B., Banerjee, S., Banerjee, N.K.: Using learned
  visual and geometric features to retrieve complete 3d proxies for broken
  objects. In: Proc. SCF. pp. 1--15. ACM, New York, NY (2021)

\bibitem{liao2018deep}
Liao, Y., Donne, S., Geiger, A.: Deep marching cubes: Learning explicit surface
  representations. In: Proc. CVPR. pp. 2916--2925. IEEE, Piscataway, NJ (2018)

\bibitem{lin2020sdf}
Lin, C.H., Wang, C., Lucey, S.: Sdf-srn: Learning signed distance 3d object
  reconstruction from static images. arXiv preprint arXiv:2010.10505
  \textbf{1}(1),  1--17 (2020)

\bibitem{lionar2021dynamic}
Lionar, S., Emtsev, D., Svilarkovic, D., Peng, S.: Dynamic plane convolutional
  occupancy networks. In: Proc.WACV. pp. 1829--1838. IEEE, Piscataway, NJ
  (2021)

\bibitem{liu2020morphing}
Liu, M., Sheng, L., Yang, S., Shao, J., Hu, S.M.: Morphing and sampling network
  for dense point cloud completion. In: Proceedings of the AAAI Conference on
  Artificial Intelligence. vol.~34, pp. 11596--11603. AAAI, Menlo Park, CA
  (2020)

\bibitem{lorensen1987marching}
Lorensen, W.E., Cline, H.E.: Marching cubes: A high resolution 3d surface
  construction algorithm. ACM SIGGRAPH Computer Graphics  \textbf{21}(4),
  163--169 (1987)

\bibitem{ma2020neural}
Ma, B., Han, Z., Liu, Y.S., Zwicker, M.: Neural-pull: Learning signed distance
  functions from point clouds by learning to pull space onto surfaces. arXiv
  preprint arXiv:2011.13495  \textbf{1}(1),  1--12 (2020)

\bibitem{mescheder2019occupancy}
Mescheder, L., Oechsle, M., Niemeyer, M., Nowozin, S., Geiger, A.: Occupancy
  networks: Learning 3d reconstruction in function space. In: Proc. CVPR. pp.
  4460--4470. IEEE, Piscataway, NJ (2019)

\bibitem{pan2021variational}
Pan, L., Chen, X., Cai, Z., Zhang, J., Zhao, H., Yi, S., Liu, Z.: Variational
  relational point completion network. In: Proceedings of the IEEE/CVF
  Conference on Computer Vision and Pattern Recognition. pp. 8524--8533 (2021)

\bibitem{papaioannou2017reassembly}
Papaioannou, G., Schreck, T., Andreadis, A., Mavridis, P., Gregor, R., Sipiran,
  I., Vardis, K.: From reassembly to object completion: A complete systems
  pipeline. Journal on Computing and Cultural Heritage  \textbf{10}(2),  1--22
  (2017)

\bibitem{park2019deepsdf}
Park, J.J., Florence, P., Straub, J., Newcombe, R., Lovegrove, S.: Deepsdf:
  Learning continuous signed distance functions for shape representation. In:
  Proc. CVPR. pp. 165--174. IEEE, Piscataway, NJ (2019)

\bibitem{peng2020convolutional}
Peng, S., Niemeyer, M., Mescheder, L., Pollefeys, M., Geiger, A.: Convolutional
  occupancy networks. In: Computer Vision--ECCV 2020: 16th European Conference,
  Glasgow, UK, August 23--28, 2020, Proceedings, Part III 16. pp. 523--540.
  Springer, Berlin, Germany (2020)

\bibitem{poursaeed2020coupling}
Poursaeed, O., Fisher, M., Aigerman, N., Kim, V.G.: Coupling explicit and
  implicit surface representations for generative 3d modeling. In: European
  Conference on Computer Vision. pp. 667--683. Springer (2020)

\bibitem{qi2017pointnet}
Qi, C.R., Su, H., Mo, K., Guibas, L.J.: Pointnet: Deep learning on point sets
  for 3d classification and segmentation. In: Proc. CVPR. pp. 652--660. IEEE,
  Piscataway, NJ (2017)

\bibitem{google2021scanned}
Research, G.: Google scanned objects.
  https://fuel.ignitionrobotics.org/1.0/\allowbreak
  GoogleResearch/fuel/collections/Google Scanned Objects (August 2021)

\bibitem{sarmad2019rl}
Sarmad, M., Lee, H.J., Kim, Y.M.: Rl-gan-net: A reinforcement learning agent
  controlled gan network for real-time point cloud shape completion. In: Proc.
  CVPR. pp. 5898--5907. IEEE, Piscataway, NJ (2019)

\bibitem{sharma2016vconv}
Sharma, A., Grau, O., Fritz, M.: Vconv-dae: Deep volumetric shape learning
  without object labels. In: ECCV. pp. 236--250. Springer, Berlin, Germany
  (2016)

\bibitem{sitzmann2020metasdf}
Sitzmann, V., Chan, E.R., Tucker, R., Snavely, N., Wetzstein, G.: Metasdf:
  Meta-learning signed distance functions. arXiv preprint arXiv:2006.09662
  \textbf{1}(1),  1--17 (2020)

\bibitem{smith2017improved}
Smith, E.J., Meger, D.: Improved adversarial systems for 3d object generation
  and reconstruction. In: Conference on Robot Learning. pp. 87--96. PMLR,
  Cambridge, UK (2017)

\bibitem{son2020saum}
Son, H., Kim, Y.M.: Saum: Symmetry-aware upsampling module for consistent point
  cloud completion. In: Proc. ACCV. pp. 1--17. Springer, Berlin, Germany (2020)

\bibitem{Stutz2018ARXIV}
Stutz, D., Geiger, A.: Learning 3d shape completion under weak supervision.
  CoRR  \textbf{abs/1805.07290} (2018), \url{http://arxiv.org/abs/1805.07290}

\bibitem{sulzer2022deep}
Sulzer, R., Landrieu, L., Boulch, A., Marlet, R., Vallet, B.: Deep surface
  reconstruction from point clouds with visibility information. arXiv preprint
  arXiv:2202.01810  (2022)

\bibitem{tang2021sign}
Tang, J., Lei, J., Xu, D., Ma, F., Jia, K., Zhang, L.: Sign-agnostic conet:
  Learning implicit surface reconstructions by sign-agnostic optimization of
  convolutional occupancy networks. arXiv preprint arXiv:2105.03582
  \textbf{1}(1),  1--16 (2021)

\bibitem{tretschk2020patchnets}
Tretschk, E., Tewari, A., Golyanik, V., Zollh{\"o}fer, M., Stoll, C., Theobalt,
  C.: Patchnets: Patch-based generalizable deep implicit 3d shape
  representations. In: Proc. ECCV. pp. 293--309. Springer, Berlin, Germany
  (2020)

\bibitem{wu2016learning}
Wu, J., Zhang, C., Xue, T., Freeman, W.T., Tenenbaum, J.B.: Learning a
  probabilistic latent space of object shapes via 3d generative-adversarial
  modeling. In: Proc. NeurIPS. pp. 82--90. Neural Information Processing
  Systems, San Diego, CA (2016)

\bibitem{xiao2022taylorimnet}
Xiao, Y., Xu, J., Gao, S.: Taylorimnet for fast 3d shape reconstruction based
  on implicit surface function. arXiv preprint arXiv:2201.06845  (2022)

\bibitem{xu2020ladybird}
Xu, Y., Fan, T., Yuan, Y., Singh, G.: Ladybird: Quasi-monte carlo sampling for
  deep implicit field based 3d reconstruction with symmetry. In: European
  Conference on Computer Vision. pp. 248--263. Springer (2020)

\bibitem{yan2022implicit}
Yan, S., Yang, Z., Li, H., Guan, L., Kang, H., Hua, G., Huang, Q.: Implicit
  autoencoder for point cloud self-supervised representation learning. arXiv
  preprint arXiv:2201.00785  (2022)

\bibitem{yan2022shapeformer}
Yan, X., Lin, L., Mitra, N.J., Lischinski, D., Cohen-Or, D., Huang, H.:
  Shapeformer: Transformer-based shape completion via sparse representation.
  arXiv preprint arXiv:2201.10326  (2022)

\bibitem{yang2021deep}
Yang, M., Wen, Y., Chen, W., Chen, Y., Jia, K.: Deep optimized priors for 3d
  shape modeling and reconstruction. In: Proc. CVPR. pp. 3269--3278. IEEE,
  Piscataway, NJ (2021)

\bibitem{yi2021complete}
Yi, L., Gong, B., Funkhouser, T.: Complete \& label: A domain adaptation
  approach to semantic segmentation of lidar point clouds. In: Proceedings of
  the IEEE/CVF Conference on Computer Vision and Pattern Recognition. pp.
  15363--15373 (2021)

\bibitem{yu2022part}
Yu, Q., Yang, C., Wei, H.: Part-wise atlasnet for 3d point cloud reconstruction
  from a single image. Knowledge-Based Systems p. 108395 (2022)

\bibitem{yuan2018pcn}
Yuan, W., Khot, T., Held, D., Mertz, C., Hebert, M.: Pcn: Point completion
  network. In: 2018 International Conference on 3D Vision (3DV). pp. 728--737.
  IEEE (2018)

\bibitem{zheng2021deep}
Zheng, Z., Yu, T., Dai, Q., Liu, Y.: Deep implicit templates for 3d shape
  representation. In: Proc. CVPR. pp. 1429--1439. IEEE, Piscataway, NJ (2021)

\end{thebibliography}
\end{document}